# Variations of images to increase their visibility

Amelia Carolina Sparavigna
Department of Applied Science and Technology
Politecnico di Torino, Torino, Italy

The calculus of variations applied to the image processing requires some numerical models able to perform the variations of images and the extremization of appropriate actions. To produce the variations of images, there are several possibilities based on the brightness maps. Before a numerical model, I propose an experimental approach, based on a tool of Gimp, GNU Image Manipulation Program, in order to visualize how the image variations can be. After the discussion of this tool, which is able to strongly increase the visibility of images, the variations and a possible functional for the visibility are proposed in the framework of a numerical model. The visibility functional is analogous to the fringe visibility of the optical interference.

**Introduction**
Sometimes, the processing of images is performed to have their optimization in some way. This is the same aim of the variational calculus. This is a calculus dealing with the problem of finding the extremal values of functionals. As maps from a set of functions to the real numbers, the functionals are often defined as definite integrals of functions and their derivatives. If we consider an image, it is possible to have several functions and functionals from it. For instance, to each pixel at the arbitrary point $P(x,y)$ in the image frame we can associate a grey or colour tone $b$ ranging from a value of 0 to 255: $b(x,y)$ is then a 2-dimensional function representative of the image intensity (brightness) distribution. From the brightness function, we can have the histogram, giving the number of pixels having the specific tone $b$. Once the functions have been defined, it is possible to devise a functional and associate the calculus of variations to the image processing [1].
Before discussing the image processing, let us remember some well-known problems of the calculus of variation. One problem is that of finding the curve of shortest length connecting two points. In a flat space, the solution is a straight line between the points. If we are in a curved space, there is the possibly of different solutions, known as geodesics. Another example is the Fermat's principle, which tells that the light follows the path of shortest optical length connecting two points.
In mechanics, this principle is known as the principle of least action. The action is a functional having as argument the trajectory, also called "path" of the system, and a real number as its result [2]. Generally, the action takes different values for different paths. Classical mechanics postulates that the path actually followed by a physical object is that minimizing the action. Figure 1 adapts an image of the Feynman's book [2]. The equations of motion of a physical system are derived from the principle of least action, by means of the variational calculus. Each path, $x(t)$, from the initial to the final position has its action. We change the path of a small amount, the variation. The new path has a different value of its action, but it is that having the lowest action, that becomes the true trajectory of the body.

In image processing, the calculus of variations is used for the image reconstruction, for instance denoising and deblurring, and for segmentations [1]. The calculus then requires some numerical models able to create the variations of images and then the minimization of some appropriate actions. To produce the variations of images, there are several possibilities based on the brightness map of the image. Before a numerical calculus, I propose an experimental approach, based on a tool of Gimp, GNU Image Manipulation Program [3], in order to visualize the image variations. After the discussion of this tool, which is able to increase strongly the visibility of images, the variations and a possible functional for the visibility are discussed with a numerical model.

**The "Curves" tool.**
Let us consider the simple case where the variation of the image is obtained by changing its brightness (more complex variations could be obtained changing shape and position of the objects in the image frame, but this is beyond the aim of this paper). To visualize how the variation can be implemented, let us use the "Curves" tool of the Gimp. "Curves" is used to control brightness, contrast and colour balance quite freely. To open this tool, the path is the following: <Image> Tools > Colour Tools > Curves. A window such as that in the right part of Fig.2 appears. As we open Curves, we see a panel reporting the histogram of the image and a straight line. Note that $x$- and $y$- axes have the grey tones as scales. The curve defines how to change the intensity of the image, with respect the diagonal that means the image as it is. Dragging the curve above the diagonal, the intensity is increased; dragging below, the intensity decreased. It is then possible to adjust the channel "Value", which will brighten or darken the image, moving the line up or down. It is also possible to adjust the Red, Blue, Green, and Alpha channels, separately. The curve is also able to redistribute contrast, because a steeper curve gives more contrast, and flat curve ranges look duller. Fig.2 shows an example of the curve manipulation. The original image has only four tones: black, white and two greys. Note the variation of the image on the left, because of the variation of the curve, shown on the right. These curves look like the paths in Fig.1.
Let us consider an example of the use of this tool of Gimp, based on the images of two geoglyphs in the Nazca Desert, Peru (Fig.3). The Nazca geoglyphs, known as the Nazca Lines, are the most famous negative geoglyphs of Peru [4]. Included among the UNESCO World Heritage Sites in 1994, the Lines are located in the Nazca Desert, a large region between the towns of Nazca and Palpa. The figures range in size up to several hundred meters and therefore can be better recognized from the air or in satellite imagery. On the Nazca plane, the geoglyphs were made by removing the uppermost surface, exposing the underlying ground, which has a different colour. This technique produces a "negative" geoglyph.
The original images, obtained from the Google Maps (see Fig.3), are grey and flat, but adjusting and redistributing the contrast, the geoglyphs becomes quite visible. Therefore, these images show an interesting application of the tool of Gimp in archaeological investigations, in particular for the processing of the satellite images. As demonstrated by several discoveries, the satellites are becoming a fundamental tool for archaeological survey. Sometimes, even the freely available images, such as those of Google Maps, directly or after a suitable processing, can provide some specific information on archaeological remains [5-7].

**Increasing the visibility**
After we have seen how variations of images can be obtained with the curves of the Gimp tool, let us try to define a functional of these curves. That is, we consider the curves of Gimp as the analogous of the paths in Fig.1. Let us indicate them with the index γ.
Consider a curve γ as obtained from the panel of the tool: the brightness of the original image changes accordingly and then

$$b(x,y) \to b_\gamma(x,y) \qquad (1)$$

For each curve γ, we have a brightness distribution $b_\gamma(x,y)$.
The average of this brightness is given by:

$$\bar{b}_\gamma = \frac{\sum_{i=1}^{N_x}\sum_{j=1}^{N_y} b_\gamma(x,y)}{N_x N_y} \qquad (2)$$

where $N_x, N_y$ are the dimensions of the image.
We can define the variance as a functional in the following way:

$$V_\gamma = \frac{\sum_{i=1}^{N_x}\sum_{j=1}^{N_y} [b_\gamma(x,y) - \bar{b}_\gamma]^2}{N_x N_y} \qquad (3)$$

$V_\gamma$ depends on curve γ. This is a functional because Eq.(3) provides a map from curve γ to a real number.
To see the behaviour of this functional as γ is varied, let us use again an image of another Nazca geoglyph. The original aerial image, quite grey ad dull is shown at the upper left corner of Fig.4. This is the geoglyph of the Condor, recorded by Raymond Ostertag [8]. The value of the functional is the number reported in the corresponding image. Dragging the curve above or below the diagonal line simply brighten or darken the image: the output images have smaller values $V_\gamma$. If a flexural point is created in the curve, as shown in the panel at the upper right corner, $V_\gamma$ increases with respect of that of the original image. Moreover, the visibility of the image is strongly increased.
The mean value (2) and variance (3) are statistical quantities, that are extensively used in parameterise the images, for applications in several research fields. For instance, as discussed in [9-11], a study of these parameters applied to the textural changes in the images of liquid crystalline cells, allows determining new phases and transitions.

**To define the visibility**
As shown by Fig.4, the functional $V_\gamma$ increases as the visibility of the geoglyph increases. Of course, a high variance $V_\gamma$ is not always corresponding to an increase of visibility. Let us imagine, for instance, changing a grey-tone image in a black and white one: the variance is strongly increased but a large part of details cancelled, and, therefore, their visibility. We know very well that increasing the contrast of images too much, means that several details are lost.

We have then to try to find another possible functional which can be a measure of the visibility of details. Here I propose a functional modelled on the problem of the visibility of interference fringes [12]. This physical quantity is also known as the "interference visibility" or "fringe visibility".

The interference of waves produces a pattern of the intensities, created by bright and dark fringes. Supposing $I_{max}$ the intensity of the bright and $I_{min}$ that of the dark fringe, the visibility is defined as:

$$V = \frac{I_{max} - I_{min}}{I_{max} + I_{min}} \qquad (4)$$

To use this visibility in the image processing, we can define the analogue of the quantities $I_{max}$, $I_{min}$ in the following manner. We consider the mean value as defined in (2), to separate the pixels in two subsets: one set contains the pixels having brightness greater or equal the mean value, the other has the pixels with $b_\gamma$ less or equal the mean value. Each set has numbers $N_\geq, N_\leq$ of pixels, respectively.

For each subset, the following mean values are evaluated:

$$\bar{b}_{min,\gamma} = \frac{1}{N_\leq}\left[\sum_{i=1}^{N_x}\sum_{j=1}^{N_y} b_\gamma(x,y)\right]_{b_\gamma \leq \bar{b}_\gamma} \quad ; \quad \bar{b}_{max,\gamma} = \frac{1}{N_\geq}\left[\sum_{i=1}^{N_x}\sum_{j=1}^{N_y} b_\gamma(x,y)\right]_{b_\gamma \geq \bar{b}_\gamma} \qquad (5)$$

Therefore, the analogous to the intensities are two mean values of the two subsets of the image pixels. In (5), the dependence on γ is explicitly written.

Let us define the visibility, analogous to the fringe visibility, as:

$$V_\gamma = \frac{\bar{b}_{max,\gamma} - \bar{b}_{min,\gamma}}{\bar{b}_{max,\gamma} + \bar{b}_{min,\gamma}} \qquad (6)$$

Assuming this $V_\gamma$ as a functional measuring of the image contrast, and therefore of its visibility, we could try to arrange some calculations in order to maximize it. We could for instance assume a theoretical variation of the brightness of the grey tones in the following manner:

$$\begin{aligned} b_\gamma &= b - a_1 \left|(b-0)\cdot(b-\bar{b})\right|^\alpha & b \leq \bar{b} \\ b_\gamma &= b + a_2 \left|(b-255)\cdot(b-\bar{b})\right|^\beta & b \geq \bar{b} \end{aligned} \qquad (7)$$

where $b$ are the brightness of the original image, and find the parameters $a_1$, $a_2$, α and β, which maximize (6). These four parameters can span some suitable ranges. In the numerical calculation, I avoided those parameters giving brightness $b_\gamma$ greater than 255 or lower than 0. α and β range from 0.1 to 1.0, $a_1$ from 0 to 5 and $a_2$ from 0 to 3.

Let me show you how the variations (7) and the extremization (maximization) of (6) work on the image of the Nazca geoglyph used for Fig.4.

Using a grey-tone version, we can have the image in Fig.5. The parameters of the variation giving the maximum value of visibility (6) are $\alpha=0.3$, $\beta=0.5$, $a_1=4.5$, and $a_2 = 1.2$ . This result is limited to the variations having brightness in the range 0,255. Note that, with respect to the original image (see Fig.4 and [8]), the visibility of the details is strongly increased.

**Conclusion**

Of course, other functions instead of (7) can be devised and used to create the image variations, and other functionals can be proposed to measure the image visibility. Future papers will be devoted to the discussion of functions and functionals. Here, the aim of the paper was that of illustrating the calculus of variations with an experimental approach. For this approach, I used a tool of Gimp, in order to visualize how the image variations can be. After the discussion of this tool, which is able to strongly increase the visibility of images (quite interesting for archaeological survey), I have proposed a visibility functional analogous to the fringe visibility of optical interference. This approach seems to give good results.

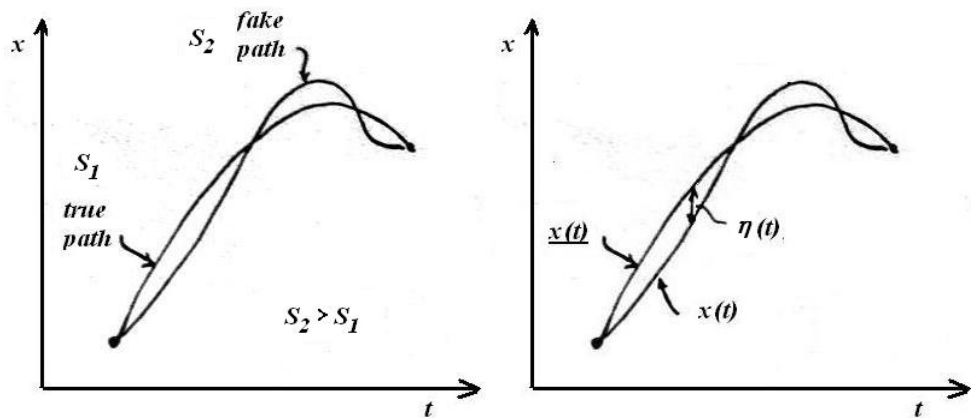

Fig.1 This image, adapted from the Feynman Lectures on Physics, illustrates the variational calculus in mechanics. To find the path, actually followed by the system, we have to calculate the action for several variations of the path. The true path is that minimizing the action $S$.

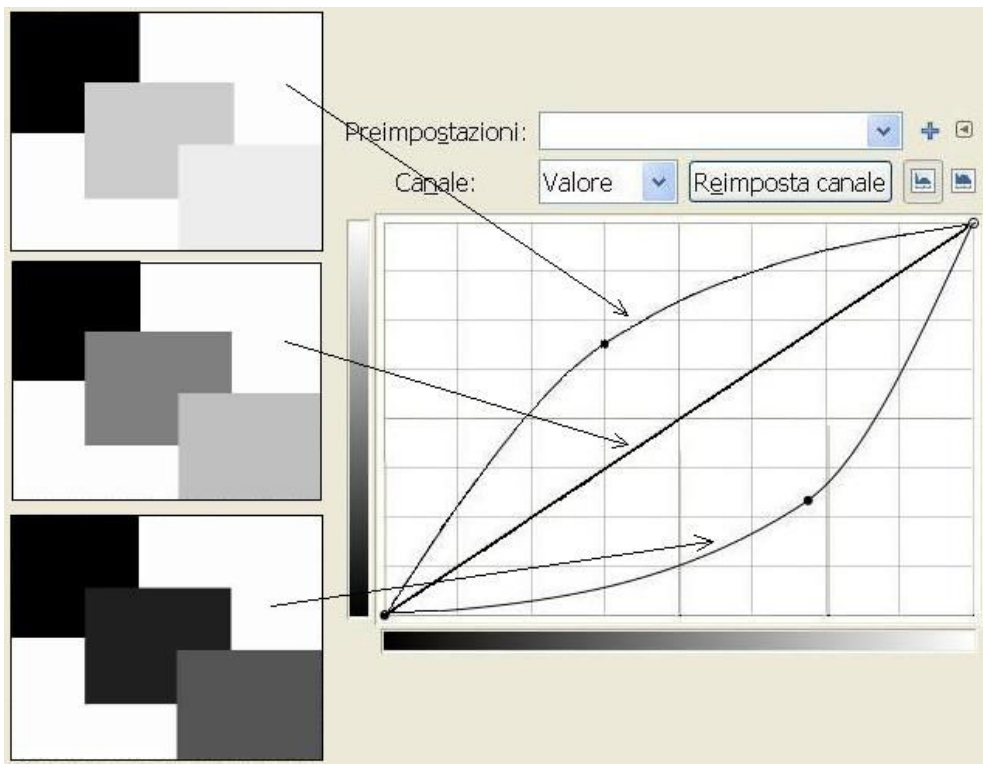

Fig.2: The "Curves" tool changes the image. Dragging the curve above the diagonal, which corresponds to the original image, the intensity is increased; dragging below, the intensity decreased. The original image has only four tones: black, white and two greys. Note the variation of the image on the left because of the variation of the curve, shown on the right. These curves look like the graphs in Fig.1.

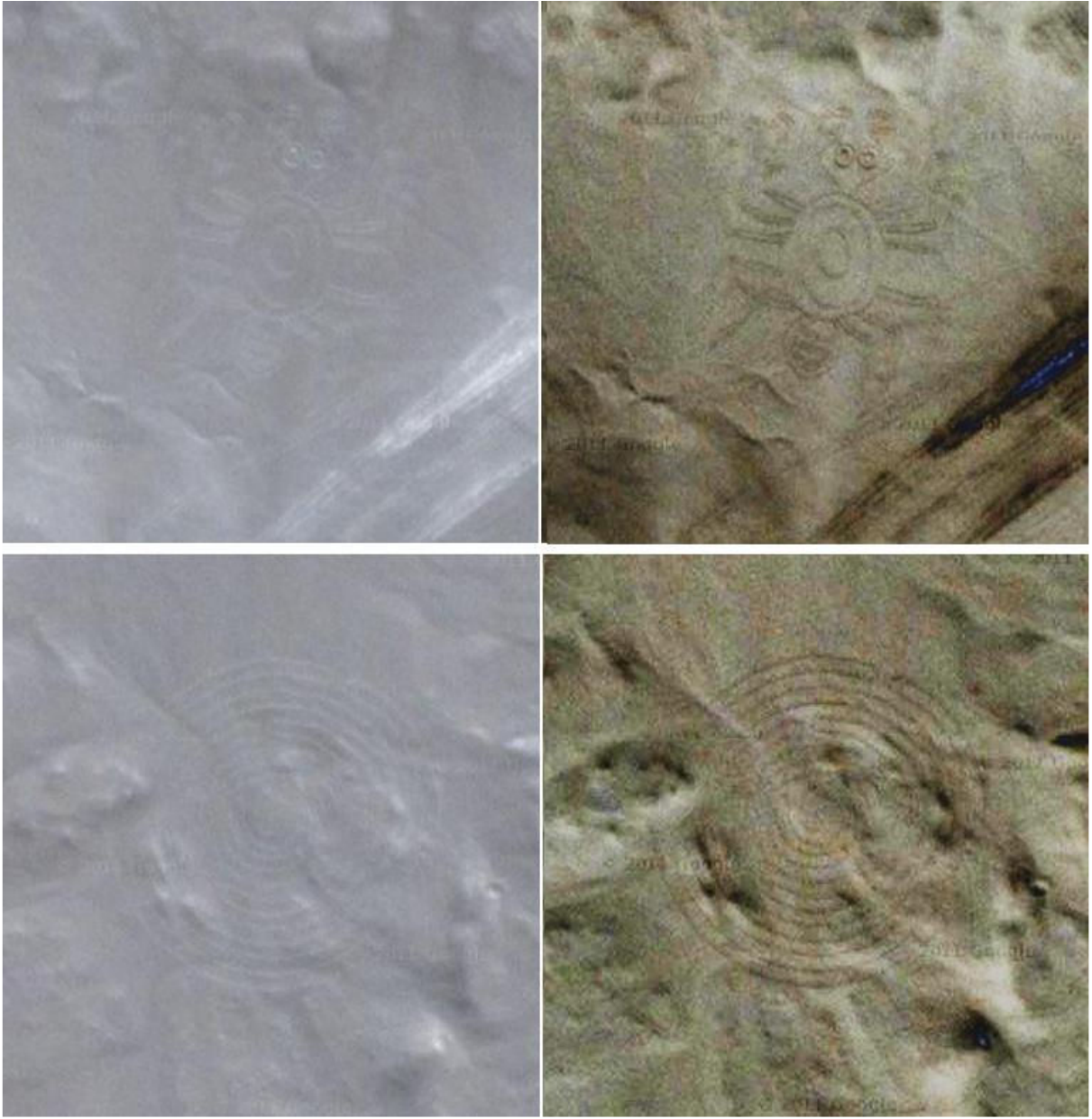

Fig. 3. Here two geoglyphs in the Nazca Desert, Peru (coordinates: 14.9997S, 75.0125W). The Nazca geoglyphs, known as the Nazca Lines, are the most famous negative geoglyphs of Peru. Included among the UNESCO World Heritage Sites in 1994, the Lines are located in the Nazca Desert, a large region between the towns of Nazca and Palpa. On the Nazca plane, the geoglyphs had been made by removing the uppermost surface, exposing the underlying ground, which has a different colour. This technique produces a "negative" geoglyph. The images on the left are as we can see in the Google Maps. On the right, the images after using the GIMP tool to change the local contrast. These images show an interesting application of this tool in archaeological investigations.

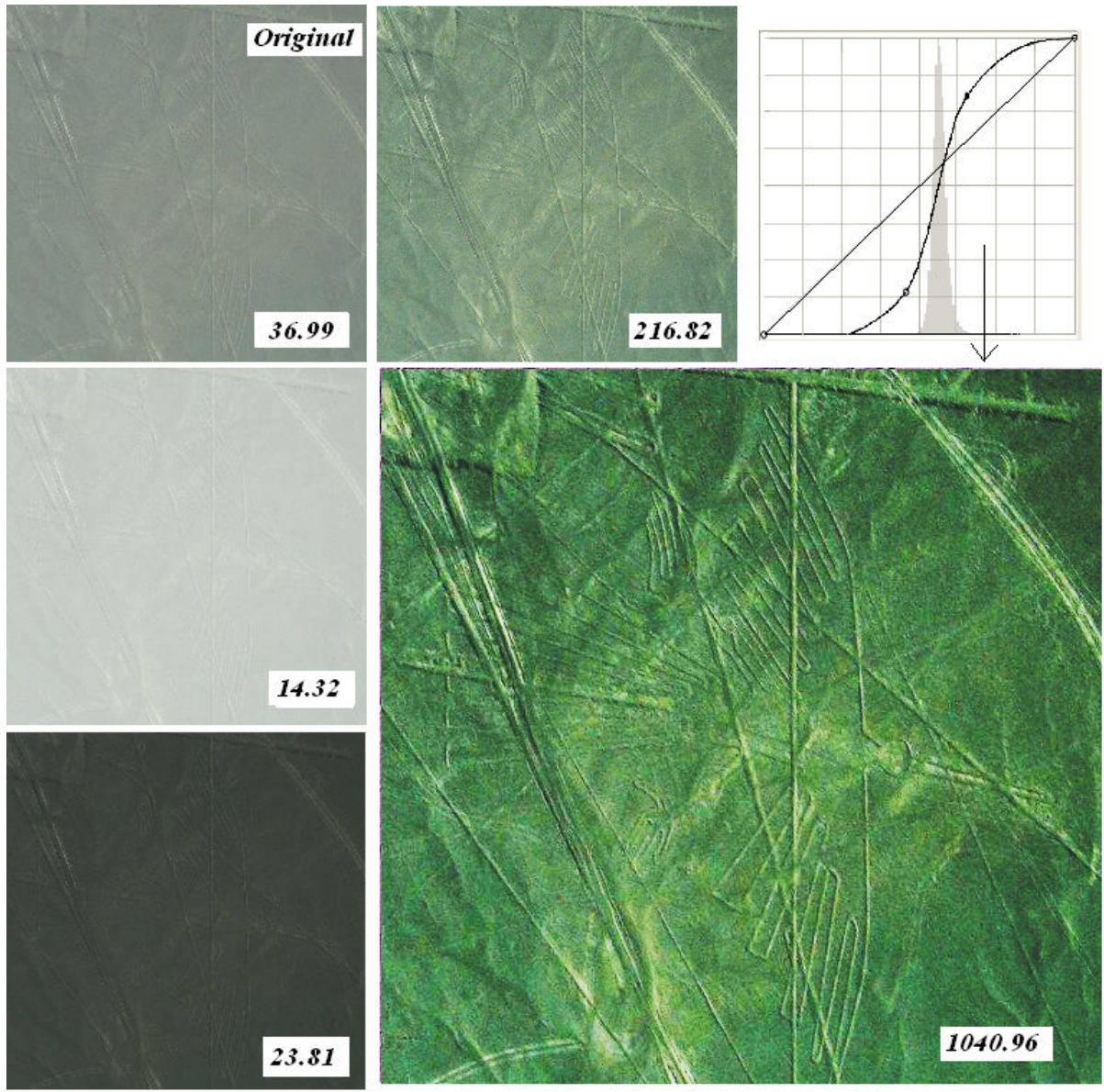

Fig.4. Let us use another Nazca geoglyph. The original aerial image, quite grey ad dull, is shown at the upper left corner. This is the geoglyph of the Condor, recoded by Raymond Ostertag. The value of the functional is the number reported in the corresponding image. Dragging the curve above or below the diagonal line produces bright or dark images: the output images have smaller values $V_\gamma$. If a flexural point is created in γ (as in the panel shown at the upper right corner), $V_\gamma$ increases with respect that of the original image. The visibility of the image is strongly increased too.

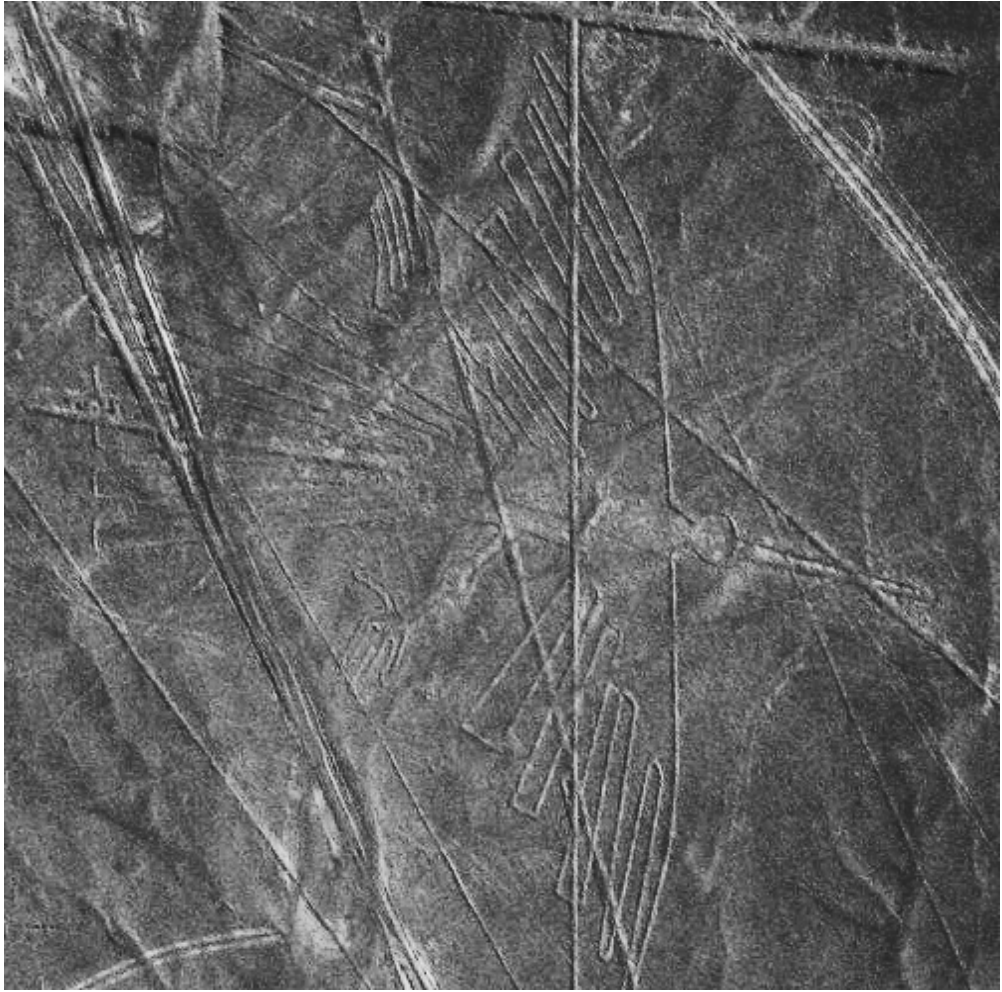

Fig.5. This is the image of the Nazca geoglyph used for Fig4, from the original aerial image recoded by Raymond Ostertag, after maximizing the visibility given by Eq.6. With respect to the original image (see Fig.4 and [8]), the visibility of details is strongly increased.